\documentclass[letterpaper]{article} 
\usepackage[]{aaai24}  
\usepackage{times}  
\usepackage{helvet}  
\usepackage{courier}  
\usepackage[hyphens]{url}  
\usepackage{graphicx} 
\usepackage{natbib} 
\usepackage{caption}
\usepackage{algorithm}
\usepackage{algorithmic}

\usepackage{newfloat}
\usepackage{listings}
\DeclareCaptionStyle{ruled}{labelfont=normalfont,labelsep=colon,strut=off} 
\floatstyle{ruled}
\newfloat{listing}{tb}{lst}{}
\floatname{listing}{Listing}

\usepackage{adjustbox}
\usepackage{booktabs}
\usepackage{subcaption}
\usepackage{url}

\newcommand{\Connections}[0]{\emph{Connections}}
\newcommand{\NYT}[0]{\emph{The New York Times}}
\newcommand{\Times}[0]{\emph{Times}}
\usepackage{xcolor}

\definecolor{yellow-c}{HTML}{FBD400}
\definecolor{green-c}{HTML}{69e352}
\definecolor{blue-c}{HTML}{5492ff}
\definecolor{purple-c}{HTML}{df7bea}

\title{Making New Connections: LLMs as Puzzle Generators for The New York Times' Connections Word Game}
\author{
    Tim Merino\textsuperscript{\rm 1},
    Sam Earle\textsuperscript{\rm 1},
    Ryan Sudhakaran\textsuperscript{\rm 2},
    Shyam Sudhakaran\textsuperscript{\rm 2},
    Julian Togelius\textsuperscript{\rm 1}
}
\affiliations{
    \textsuperscript{\rm 1}New York University,\\
    \textsuperscript{\rm 2}Jester Labs,\\
    tm3477@nyu.edu, sam.earle@nyu.edu, ryan@jesterlabs.ai, shyam@jesterlabs.ai, julian@togelius.com
}

\begin{document}
\pagenumbering{arabic}

\maketitle

\begin{abstract}
The \Connections{} puzzle is a word association game published daily by \textit{The New York Times} (NYT). In this game, players are asked to find groups of four words that are connected by a common theme. While solving a given Connections puzzle requires both semantic knowledge and abstract reasoning, generating novel puzzles additionally requires a form of metacognition: generators must be able to accurately model the downstream reasoning of potential solvers. In this paper, we investigate the ability of the GPT family of Large Language Models (LLMs) to generate challenging and creative word games for human players. We start with an analysis of the word game \Connections{} and the unique challenges it poses as a Procedural Content Generation (PCG) domain. We then propose a method for generating \Connections{} puzzles using LLMs by adapting a Tree of Thoughts (ToT) prompting approach. We evaluate this method by conducting a user study, asking human players to compare AI-generated puzzles against published \Connections{} puzzles. Our findings show that LLMs are capable puzzle creators, and can generate diverse sets of enjoyable, challenging, and creative \Connections{} puzzles as judged by human users. 
\end{abstract}

\section{Introduction}

Word games, in which game mechanics revolve around manipulating, parsing, and/or inventing words, have been around for a very long time. While crosswords as we know them today have only been around for just over a century, word puzzles have existed at least since the Romans~\cite{raphael20brief}. More recently, online word games have seen increased popularity, including several novel word games published by the New York Times. While some research has been done on solving human-authored word games, the challenge of generating novel word games has seen scant attention from AI researchers.

In this paper, we study the popular word game \Connections{}. This is a game about semantic clustering: the player is given 16 words and must sort them into four categories. The apparent simplicity of this game belies a significant depth owing to the large numbers of different organizing principles for these categories, as well as various deceptive strategies in generating these puzzles. 

In this work, we present a method for generating and evaluating \Connections{} puzzles using LLMs and validate our approach with a user study that compares AI generated puzzles against those authored by humans.

The impact of our work is manifold. First, we provide a new source of puzzles for a game that, at time of writing, has only 380 published puzzles. In addition, our prompting approach provides a framework that can be applied to other word games, paving the way for new applications in procedural content generation. Our study of puzzle generation methods also shines light on the design of the \Connections{} puzzle itself---through our user study we can observe the effects of various puzzle-generating strategies or heuristics on the end user experience. Finally, through our experiments we gain some insight into the capabilities and weaknesses of the underlying large language models and highlight valuable avenues for future work.\footnote{We make our code, the dataset of \Times{} puzzles, our generated puzzles, and full set of prompts available at \url{https://anonymous.4open.science/r/making-new-connections-78D1}}

\section{Related Work}

\subsection{LLMs for games}

While LLMs are a relatively recent invention, there is already burgeoning literature on applying LLMs to games, as catalogued by~\cite{gallotta2024large}. This includes using LLMs to play games, generate games, generate dialogue, implement game mechanics, and more.

Several studies have investigated the application of LLMs to level generation. MarioGPT~\cite{sudhakaran2023prompt} fine-tunes GPT-2 to generate levels in the side-scrolling platformer \textit{Super Mario Bros.}, conditioned on target characteristics of the generated level such as number of jumps, coins and enemies. Similarly, \citeauthor{todd2023level} fine-tune GPT-2 and GPT-3 to generate Sokoban levels. In both cases, the resulting models are capable of generating novel and playable levels.

These prior LLM-based level-generation works above focus on embodied grid-worlds. Though these representations can be thought of as existing at a sub-linguistic level, results seem to suggest that pre-training on vast amount of natural language data seems to nonetheless result in useful priors for this task. In this paper, we investigate generating ``levels'' in a game that operates instead directly at the level of language. Unlike in grid-worlds, training a model from scratch strictly on human-generated levels would be a virtually infeasible approach (due to the sparsity of level data relative to the vast semantic complexity of human language), and pre-training on a vast corpus of human knowledge is a necessary pre-requisite for generating any novel puzzles at all.

Though generation of \Connections{} puzzles is an unexplored domain, recent research has studied \textit{solving} language-based games. \citeauthor{Jaramillo_Charity_Canaan_Togelius_2020} implement and evaluate various AI agents for the social board game \textit{Codenames}. \textit{Codenames}, similar to \Connections{}, uses semantic clustering of seemingly unrelated words as a core game mechanic. They find that a GPT-2 word embedding based bot outperforms non-vector approaches such as TF-IDF and Naive-Bayes. We continue this exploration of Language Models for word association games, utilizing the latest GPT models.

Recent work has studied LLMs in the domain of \Connections{}. \citet{todd2024missed} first study the task of solving NYT Connections puzzles, finding that GPT-4 is able to solve $38.93\%$ of puzzles using Chain of Thought prompting. They find that sentence embedding models, using a cosine similarity guessing strategy, are able to solve $11.6\%$ of puzzles, outperforming the GPT-3.5 model. Later work further explores \Connections{} solving via LLMs, categorizing the knowledge types required to solve the puzzle and evaluating a variety of LLMs on a modified version of the solving task \cite{connectingdots}. The best performing model, GPT-4o, is only able to solve $8\%$ of puzzles. They note that certain ``reasoning types'' present in categories are significantly harder for LLMs to solve, highlighting the diversity of semantic reasoning inherent to the game.

\subsection{The Connections Puzzle}
\begin{figure}[t]
    \centering
    \includegraphics[width=\columnwidth,trim={0 20 0 0}]{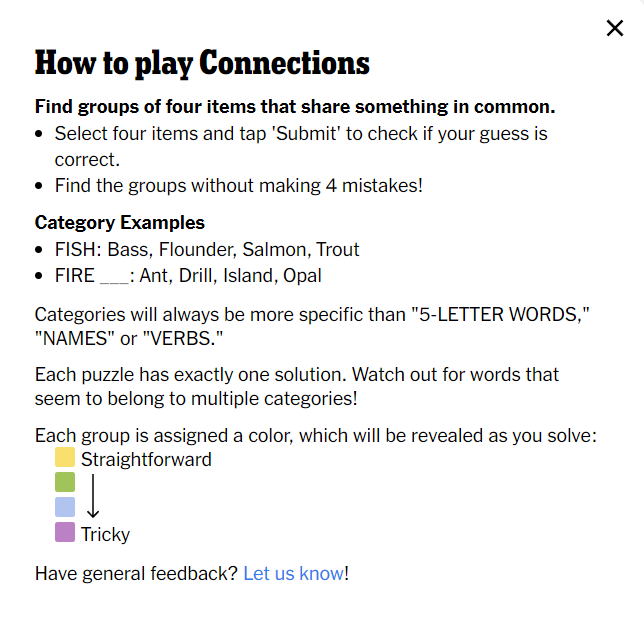}
    \caption{The help page of the \Connections{} puzzle}
    \label{fig:connections_puzzle}
\end{figure}

\begin{figure}[b]
    \centering
    \includegraphics[width=\columnwidth,trim={0 0 0 0}]{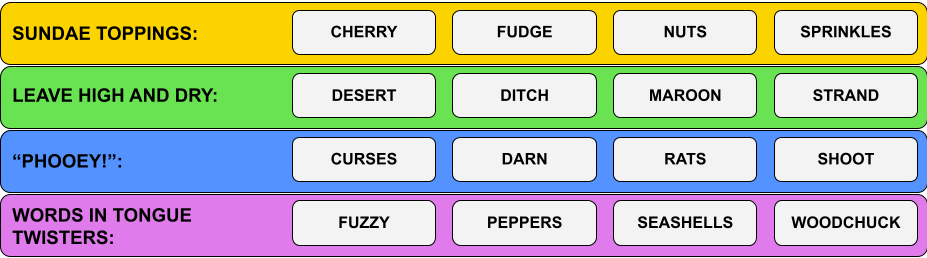}
    \caption{A \Connections{} puzzle, published by The New York Times on Jan 19, 2024}
    \label{fig:nyt_example}
\end{figure}
\Connections{} is a category-matching word game published daily by The New York Times. The game debuted on June 12, 2023, and quickly became one of their most played games. 
A \Connections{} game board is represented by a grid of sixteen English words or names. The arrangement of words in the grid is semi-random\footnote{Some puzzles have intentionally spelled out phrases in the first row of words, though this phrase is never a word group} and can be arbitrarily re-shuffled by the player. In order to submit a guess, the player selects a set of four words they believe are connected under a common theme. If the guess is correct (i.e. the four words belong to the same category) then the category is revealed and the words are removed from the grid. If three of the four words belong to the same category, this fact is indicated to the player with the message ``One away!'', but neither the words nor the category are revealed. If two or fewer words belong to the same category then no additional information is provided to the player. If the player makes four incorrect guesses, the puzzle is failed.

Each category within a puzzle has an assigned color which represent the its difficulty. In order of ascending difficulty, these colors are: yellow (most straightforward), green, blue, and purple (trickiest). Straightforward categories often use the most common definition of words. An example given by the tutorial is the category \textit{FISH}, containing ``bass", ``flounder", ``salmon", and ``trout". For trickier groups, the connection between words can be much more complex. Purple groups may use rhymes, homophones, anagrams, or ``fill in the blank" clues such as \textit{FIRE \_\_\_}\cite{Aronow_Levine_2023}.

Wyna Liu, who creates the daily game boards for \textit{Connections}, has discussed some aspects of puzzle creation in articles published by The Times~\cite{Liu_nyt}. \Connections{} puzzles are created in three steps: First, a puzzle creator compiles a candidate puzzle. This puzzle is given to a puzzle editor, who tests and edits the game board. Finally, a panel of testers evaluate the quality and challenge level of the puzzle.

For a puzzle to feel challenging and satisfying, Liu notes there has to be a mix of category ``types", such as wordplay categories and synonym categories. She cites three axes along which difficulty may be modulated: familiarity of words, ambiguity of their categorization, and variety of the wordplay~\cite{Liu_nyt}.

Generation of puzzles that match both the challenge and satisfaction of those published by The New York Times requires both creativity, abstract reasoning, and deep semantic understanding. Categories must have a range of difficulties, and incorporate abstract connections. Words must logically fit into a category, without being too obvious for the player. As a whole, the words and categories of a puzzle must play off each other, with interplay between words to trick the player. Satisfying all of these constraints makes \Connections{} a unique challenge for modern LLM systems. 

\subsection{Prompting}

Because LLMs are trained auto-regressively to fill in or complete sequences of tokens from the training data, their output at inference time depends entirely on the ``prompt'' which they are provided---i.e. the piece of text which they are effectively asked to complete.

As a result, LLM performance on downstream tasks can be strongly influenced by the technique used to generate the prompt.

A vast amount of research has studied how to most effectively prompt language models~\cite{schulhoff2024prompt}. In Chain-of-Thought~\cite{wei2022chain}, for example, an LLM is asked to show its reasoning step by step. Results suggest that by mimicking a more deliberate and thorough thought process, the LLM is less likely to jump to incorrect conclusions, and more inclined toward coherent reasoning. In its simplest form, Chain-of-Thought amounts to appending the phrase ``let's think step-by-step'' to the prompt. 

In the Tree-of-Thought prompting approach (ToT)~\cite{yao2023tree}, tasks are explicitly broken down into sub-tasks by the human user. Each sub-task is considered as a level in a tree, and the LLM is queried multiple times (while varying random seed or other hyperparameters) to produce multiple child nodes at each level.
Then, some evaluation metric is calculated for each leaf-node in the tree, and tree nodes are expanded greedily, prioritising higher-scoring nodes.
ToT has been shown to improve performance on challenging tasks such as the Game of 24~\cite{yao2023tree} and Sodoku~\cite{long2023large}, where simpler prompting techniques such as Chain of Thought~\cite{wei2022chain} fail.
In these tasks, there are verifiable solutions to the problem, allowing for the explicit scoring and ranking of candidate solutions generated by an LLM. In more open-ended tasks, where success is much more difficult to define, there is no simple way to measure the quality of an LLM generator's output.

One such task is the creative writing task explored by \citet{yao2023tree}. They generate 4 random sentences, and task the LLM with generating a four paragraph passage where each paragraph ends with one of the random sentences. While there is a simple constraint that can be checked (whether each paragraph ends with the sentence provided), evaluating the quality of the writing as a whole is much more complex. Separate LLM agents can be used to evaluate and rate these outputs, but existing research shows they may be biased towards LLM-generated text~\cite{koo2023benchmarking}.

\section{Methodology}

Our generative pipeline is loosely inspired by the creative process used by the New York Times and consists of three components: a puzzle \textit{creator}, a puzzle \textit{editor}, and a human puzzle \textit{evaluator}. GPT-4 serves as the puzzle creator, generating sets of puzzles using an iterative approach loosely based on the Tree of Thoughts algorithm. This set of generated puzzles is passed to the puzzle editor---a separate LLM instance that can make minor changes to the category names. Next, the set of edited puzzles is manually evaluated to determine the ``best'' puzzles. We evaluate the quality of our generated puzzles by testing the evaluator's top puzzles in a user study.

\subsection{Identifying Puzzle Constraints}
\label{sec:constraints}
Unlike their Crossword puzzle~\cite{Aronow_crossword}, \NYT{} does not publish rules or guidelines for creating a \Connections{} game board. We start by identifying constraints that define valid \Connections{} puzzles. By looking at existing puzzles and the in-game tutorial (Fig \ref{fig:connections_puzzle}), we infer some basic constraints that a \Connections{} puzzle must satisfy:
\begin{enumerate}
\item The puzzle contains 16 unique words.
\item The puzzle has 4 unique groups of 4 words.
\item All words in a group share a connection.
\item Each word is used exactly once.
\item The puzzle has exactly one solution.
\item Categories are more specific than 5-LETTER WORDS, NAMES, or VERBS.
\end{enumerate}

This basic set of constraints defines a large puzzle space, containing all possible combinations of 4 valid word groups. The vast majority of these puzzles are trivially easy to solve---players will notice the obvious semantic clustering of words in the puzzle, eliminating the satisfaction of playing the game. In order to generate puzzles that are challenging and fun to play, we identify general principles that are followed by the \Times{} puzzle creators. We incorporate these principles into our prompts to guide puzzle creation towards the style of puzzle published by the \Times{}. These principles include:

{\flushleft\textbf{Varied Categories:}} Categories within a given puzzle must be thematically distinct (i.e. a COLORS category will not share a puzzle with a PRIMARY COLORS category)
{\flushleft\textbf{Unique Names:}} Words in the category name can't be appear as part of that category (i.e. the word GREEN will not appear in the category SHADES OF GREEN)
{\flushleft\textbf{Spelling Matters:}} Words must belong to their group in the given spelling (i.e. BE will not appear in the category INSECT, but BEE might).

\subsection{Puzzle Difficulty}

\begin{figure*}
\begin{subfigure}[ht]{\columnwidth}
    \centering
    \includegraphics[width=\columnwidth,trim={0 0 0 0}]{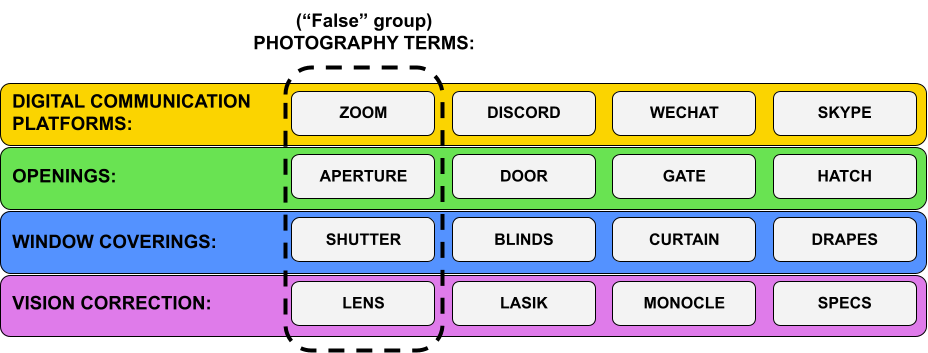}
    \caption{A generated ``False Group'' puzzle. The ``false group'' is circled in red, but is not a correct group in the puzzle's solution. The correct groups are circled in their respective color.}
    \label{fig:false_group}
\end{subfigure}
\hfill
\begin{subfigure}[ht]{\columnwidth}
    \centering
\vspace{0.4cm}
    \includegraphics[width=\columnwidth,trim={0 0 0 0}]{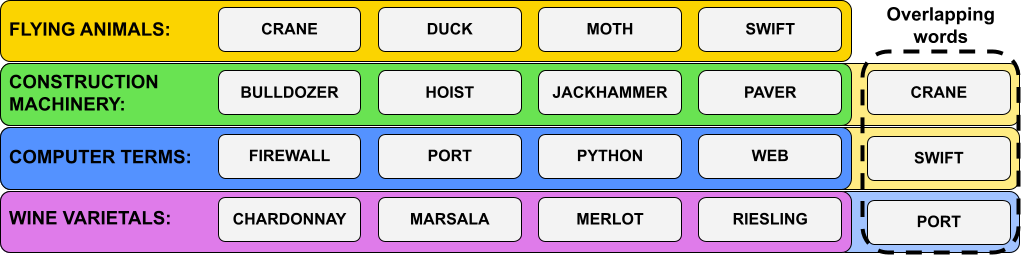}
    \caption{A generated ``Intentional Overlap'' puzzle. The correct groups are circled in their respective color. Each group after the first appears to include an extra word from a separate group.}
    \label{fig:overlap_group}
\end{subfigure}
\begin{subfigure}{\columnwidth}
    \centering
    \includegraphics[width=\columnwidth,trim={0 0 0 0}]{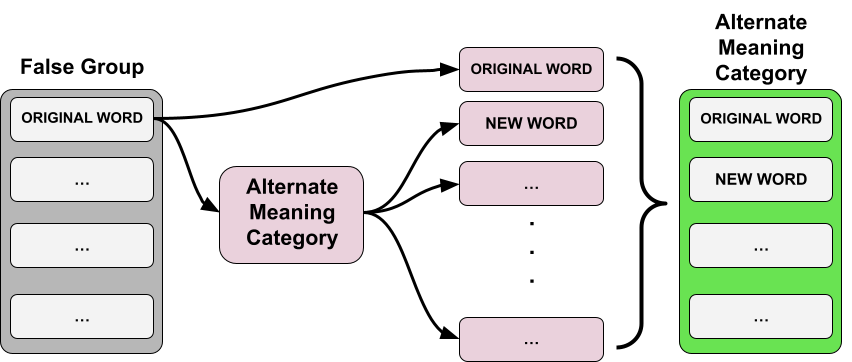}
    \caption{Illustration showing the process for generating groups in the ``False Group'' pipeline}
    \label{fig:false_group_pipeline}
\end{subfigure}
\hfill
\begin{subfigure}{\columnwidth}
    \centering
    \includegraphics[width=\columnwidth,trim={0 0 0 0}]{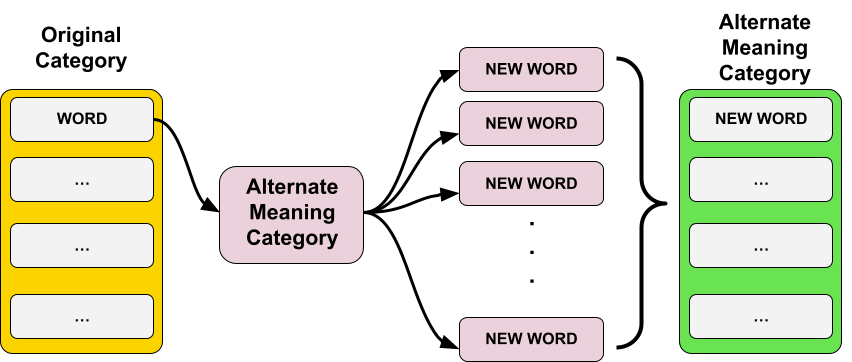}
    \caption{Illustration showing the process for generating new groups in the ``Intentional Overlap'' pipeline}
    \label{fig:overlap_pipeline}
\end{subfigure}
\caption{Overview of LLM-driven pipelines for generating \Connections{} puzzles with false or intentionally overlapping groups.}
\end{figure*}

Making engaging puzzles requires a certain level of difficulty. One aspect of this is explicitly stated in the tutorial of the game: ``Watch out for words that seem to belong to multiple categories!" (see Figure~\ref{fig:connections_puzzle}). From this, we deduce that a ``good" \Connections{} puzzle must entice the player to make mistakes by using ``misleading'' words.

We identify two distinct ways this occurs in official \textit{Connections} puzzles. The first is using ``overlap words"---words that seem to belong to multiple categories. 
An example of this can be seen in Figure \ref{fig:nyt_example}. Words like ``fudge'' and ``nuts'' seem like they could belong in the \textit{PHOOEY!} category, but actually belong to \textit{SUNDAE TOPPINGS}.
The presence of these words forces the player to shift focus and consider multiple potential groups at once in order to disambiguate the group membership of individual words. This is cited as a key feature that is sought out by the \Times{}' puzzle editors: ``More overlap makes a harder puzzle.”~\cite{Aronow_Levine_2023} 

The other form of difficulty is what we name ``false groups". These are connections between words that a player may spot, but in actuality are not related to any of the puzzle's true groups.
An example of a false connection in Figure \ref{fig:nyt_example} could be ``rats'' and ``woodchuck'' both being \textit{rodents}, or ``cherry'' and ``peppers'' both being \textit{red foods}.
Whether a puzzle appears to present a false group is ultimately as much a function of the player as it is of the puzzle itself. Nevertheless, even if these dead ends don't actually elicit an incorrect guess from the player, they arguably make the process of solving the puzzle more engaging by adding richness to the puzzle's semantic landscape.

We select overlap words and false groups as the main avenues by which to incorporate challenge into our generated puzzles.

\subsection{Puzzle Creation pipeline} \label{sec:generative_pipeline}
We decompose the task of \Connections{} puzzle generation into an iterative process: each generative step equals the creation a word group, composed of a category name along with a pool of words that belong to it. Although the puzzle could be decomposed further, with every step providing a single word or category name, this approach would become quite computationally expensive using current LLMs. We set the pool size to be eight words. We find that this results in GPT suggesting multiple valid words of varying difficulty, without suggesting illogical words in order to meet the quota. We use \texttt{gpt-4-1106-preview} via OpenAI's API as our word group creator, as well as all other LLMs in the pipeline.

We use a few-shot ``propose prompt" approach for word group generation, inspired by the prompts used in prior work for the Tree of Thoughts Crossword Solving task \cite{yao2023tree}. This approach prompts an LLM to generate one or many word groups in a single context, which are then parsed out to become individual ``nodes''. In some puzzle types, such as the ``intentional overlap'' puzzles, we generate two distinct categories and word pools in each prompt. In others, only one category is taken at each step. 
.

One issue facing GPT models is limited variety in output. In early experiments, our group-generation LLM would frequently propose the same limited set of categories and words, such as \textit{BOARD GAMES}: ``chess", ``checkers", ``monopoly", ``life". Varying seed and temperature did little to curb this behavior.

We solve this problem by injecting semi-random text into the first group-generation prompt. We add a task to the LLM's system prompt: first, write a short story using user-provided words, then use that story as inspiration for creating your categories. In the user prompt, we inject four randomly selected words from published \Connections{} puzzles. We find this prompting technique leads to much higher diversity of output, with very few repeated categories across our set of generated puzzles.

These proposed categories still lacked the challenging and abstract categories present in real \textit{Connections} puzzles, as the LLM has no knowledge of the \Connections{} puzzle itself\footnote{Per OpenAI, \texttt{gpt-4-1106-preview} has a knowledge cut-off of April 2023, before the first \Connections{} puzzle was released}. To solve this, we analyzed published \textit{Connections} puzzles and identified common category ``styles'', and collect a set of example category names. Some example styles are ``Synonyms or Slang'', ``Wordplay'', and ``Fill in the blank'' categories. We add these category styles, along with a brief description and three example category names, into all proposal prompts. The LLM is instructed to only craft categories that fit one of the given styles. This leads to more complex proposed categories, such as \textit{WORDS THAT CAN FOLLOW ``FIRE''}.

\paragraph{Intentional Overlap and False Connection trees}
We build two different styles of puzzle, which are distinct in the way the LLM introduces tricky word interplay. Both methods use the same prompt to generate the ``root" word group, but differ in followup group generation.

In intentional overlap trees, we first create groups of four words from the word pool, and include these previously generated groups as context in the next group-generation prompt. For each previously used word, the LLM proposes an alternate category using a different meaning of the word. It then selects one alternate category, and generates a new pool of eight words that could fit under it. This introduces the ``overlap" difficulty into the puzzle---each followup group appears to include one additional word from a different group. This process is illustrated in Figure \ref{fig:overlap_pipeline}

In false connection trees, the root group becomes the ``false group'', which is not directly included in the puzzle. Instead, each of the four words from the false group are used as a basis for an entirely new word group, using an alternate definition. Only these followup groups are used to form the final puzzle (See Figure \ref{fig:false_group_pipeline}). This introduces the ``false connection'' difficulty: there appears to be five valid groupings of words---the root group and four followup groups---but only the latter four are valid.

\paragraph{Creating groups of four}
Initial experiments indicate that LLMs struggle to estimate the difficulty of \Connections{} word groups, and fail to select subsets of four words in a way that aligns with human judgement. Rather than using an LLM to group words together, we use a text embedding similarity metric to groups words from the word pools created by the LLM. This provides two benefits: we eliminate a failure point from our pipeline stemming from LLM hallucinations and stochasticity, and are able to create and assign a ``difficulty'' score to all possible combinations of four words.

We use the \textsc{MPNet} embedding model from the \textit{SentenceTransformers} library, which had the top performance in the task of solving the \Connections{} puzzle \cite{todd2024missed}. We infer that the high performance in the solving task means the \textsc{MPNet} model encodes a word's semantic information in a way most similar to their usage in the \Connections{} puzzle. 

We first test our hypothesis that intra-group embedding similarity is a proxy for word group difficulty on published \Connections{} puzzles. We compute the average pairwise cosine similarity between embeddings of all words in existing puzzle groups, shown in Table \ref{tab:cossim_nyt}. We then average these score across groups of the same color. We find that the cosine similarity metric aligns with the \NYT{} difficulty scale, with yellow groups having higher similarity, and purple groups having lower similarity.

\begin{table}[t]
    \centering
    \caption{Average intra-group pairwise cosine similarity by group color across 150 \Connections{} puzzles}
    
    \begin{tabular}{lcc}
    \toprule
    \textbf{Color} &  \textbf{Average similarity score} & \textbf{Variance}\\
    \midrule
    Yellow &  0.4017 & 0.0285\\
    Green & 0.3536 & 0.0214\\
    Blue &  0.2905 & 0.0123\\
    Purple  &  0.2664 & 0.0108\\
    \bottomrule
    \end{tabular}
    \label{tab:cossim_nyt}
\end{table}

For each LLM-proposed word pool, we compute this similarity metric for all combinations of four words. We then create four groups corresponding to the color-difficulty levels. Yellow and Purple groups are the combinations with highest and lowest similarity, respectively, with green and blue groups selected as the combinations closest to $\frac{1}{3}(max\_similarity - min\_similarity)$ and $\frac{2}{3}(max\_similarity - min\_similarity)$

This method results in 24 final versions of a single puzzle---each word group has four distinct ``difficulty'' groupings, with different words but the same category.

\subsection{Puzzle editing}

Due to the stochastic nature of LLMs and their sensitivity to prompts, not all final puzzles are perfect. A common failure mode we observe is when there is a clear connection between words in a group, but the category name misrepresents that connection. An example is the generated group \textit{HAWK THE WARES}: ``wares'', ``items'', ``goods'', ``merchandise''. \textit{HAWK THE WARES} does not accurately describe how these words are connected, but \textit{THINGS FOR SALE} would. For \textit{Connections} puzzles, a small mistake like this in a word or category name can ruin the solving experience.

To increase the efficiency of our method, we use a ``Puzzle Editor" prompt.
First, it identifies the connecting theme between each word group in a complete puzzle. Then, it evaluates whether the existing category name is accurate. If not, it rewrites the category name so that it truthfully represents the connection. This often fixes otherwise invalid word groups, and rewritten category names are often much closer in style to real \textit{Connections} categories.

The final component in our generation pipeline is color-coded difficulty ranking. While we assign colors based on the similarity metric generated via our embedding model, we consider puzzles where two groups are the same ``color'' during evaluation. Additionally, this method of assigning colors does not consider the relative difficulty of other groups in the puzzle. To create a puzzle with all four colors, we pass generated puzzles to a final LLM using a ``difficulty ranking" prompt. This LLM assigns colors (yellow, green, blue, purple) to the groups of each puzzle based on the LLM's perception of their difficulty.

\section{Human Preference Experiment}

\subsection{Datasets}
To evaluate the appeal of our AI-generated Connection puzzles, we conduct a user study to evaluate human preferences between AI generated and real \Connections{} puzzles.

We create four sets of AI generated puzzles:
\begin{enumerate}
\item A baseline ``one-step'' LLM generated set
\item An ``intentional overlap'' LLM generated set
\item A ``false connection'' LLM generated set
\item A ``false connection'' LLM generated set, where the ``false group'' is taken from a real \Connections{} puzzle
\end{enumerate}

\paragraph{Baseline Set} To evaluate the effectiveness of our iterative pipeline over simpler prompting methods, we generate a set of ``One-Step`` puzzles to act as a baseline for LLM puzzle generation. In this method, we ask an LLM to generate a complete \Connections{} puzzle in a single prompt. We provide a condensed version of the prompts used in the iterative approach, as well as few-shot examples of real \Connections{} puzzles.

\paragraph{Iteratively Generated Set} We generate three sub-types of puzzle using the iterative generation strategy described in Methodology. For each sub-type, we generate 15 complete sets of puzzles, and 5 puzzles are manually selected for inclusion into the user study. We use \texttt{gpt-4-1106-preview} via OpenAI's chat completion API, with temperature 1 for all generation and editing prompts.

5 puzzles use the ``intentional overlap'' method, where each category after the first is created by using an alterate definition of a previous groups' word. 

10 puzzles are generated using the ``false group'' method. In half of these puzzles, the LLM proposes its own false group category and word pool, and the group of four words with the highest similarity metric (the ``yellow'' group) is used. We refer to these as ``LLM False Group'' puzzles. In the other half, we manually select word groups from existing \Connections{} puzzles to act as the false group. We manually select groups where each word has a clear alternate definition, so that a valid puzzle with thematically distinct categories can be generated (See Table \ref{table:seed_categories} for exact groups). 

We include this ``seeded'' set of false-group puzzles as an initial exploration into human-in-the-loop generation of \Connections{} puzzles. While fully-automated PCG systems allow for infinite artifacts with no human intervention, it is rare that such systems perform perfectly. We seek to explore how including some human creativity into the generation process, by way of real puzzle groups, affects human evaluation of the puzzles. 

\paragraph{Real set} Finally, we randomly sample $20$ previously published New York Times \Connections{} puzzles to form our real puzzle set.

\subsection{Experiment setup}
We implement a web-based replica of the \Connections{} interface, following the same rules used by the New York Times. Players are allowed four incorrect guesses before failing the puzzle. They are able to randomly shuffle the game board, and are told when their guess is ``one away'' from a correct answer. The puzzles are integrated into a survey page, where users are asked to play two \Connections{} puzzles and answer a series of questions. If the user does not successfully solve the puzzle, all categories are revealed so the user can evaluate the complete puzzles.

Pairs of \Connections{} puzzles, one AI-generated and one from the \Times{}, are randomly selected at the start of each survey. The ordering in which the two puzzles is randomized.

We collect gameplay from the user's play sessions on both puzzles, including all guesses made and whether or not they completed the puzzle. After completing both puzzles, the user is asked the following questions:
\begin{enumerate}
\item  What is your English Proficiency?
\item  How often do you play \Connections{}?
\item  Have you seen either puzzle before?
\item  Which puzzle was more creative?
\item  Which puzzle was harder?
\item  Which puzzle did you like more?
\end{enumerate}
For questions 4-6, users can select ``Puzzle 1", ``Puzzle 2", or ``Tie/Neither".
Optional free response questions are included for questions 4-6, allowing users to explain their reasoning. A final free response is included for miscellaneous comments.

To avoid potential bias in user responses, we do not explicitly state which puzzle is from which source, nor that the puzzles are AI generated. Additionally, we ask users to self-report if they have seen either puzzle before. We do not include responses where a user recognized a puzzle in our analysis.

\section{Results}

\begin{figure*}[t]
\centering
\includegraphics[width=0.8\textwidth]{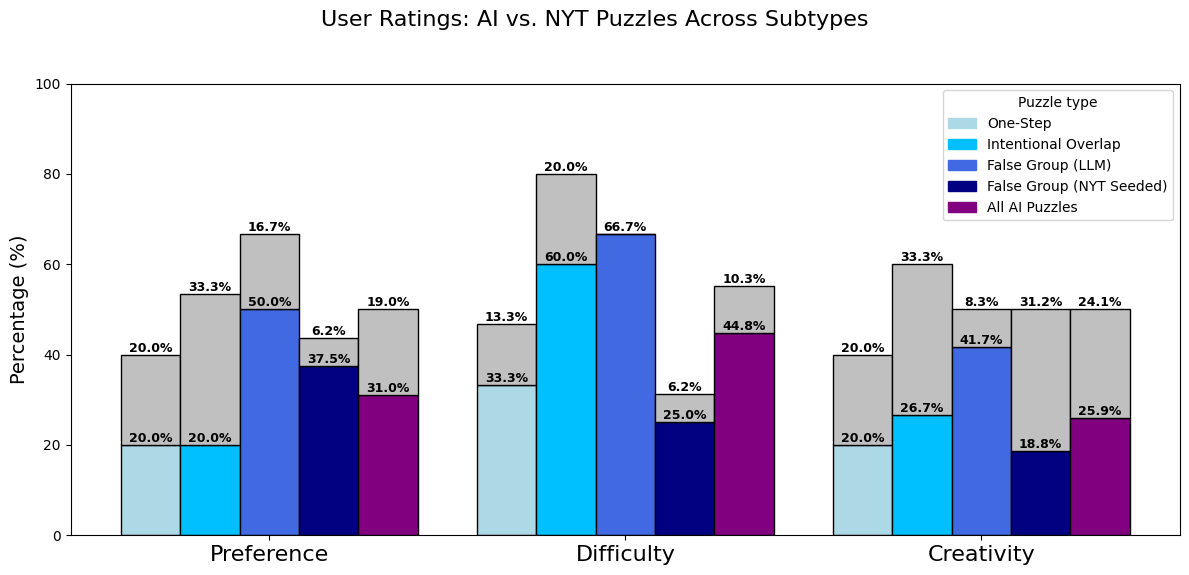}
\caption{User response data for survey questions 4-6. Colored bars represent AI puzzle preference, while gray bars represent ``Tie / Neither''. AI-generated puzzles with LLM-generated false groups, in particular, are competitive with NYT puzzles in terms of user preference and perceived creativity, while being judged generally more difficult.}
\label{preference_data}
\end{figure*}

\begin{figure*}[t]
\centering
\includegraphics[width=0.8\textwidth]{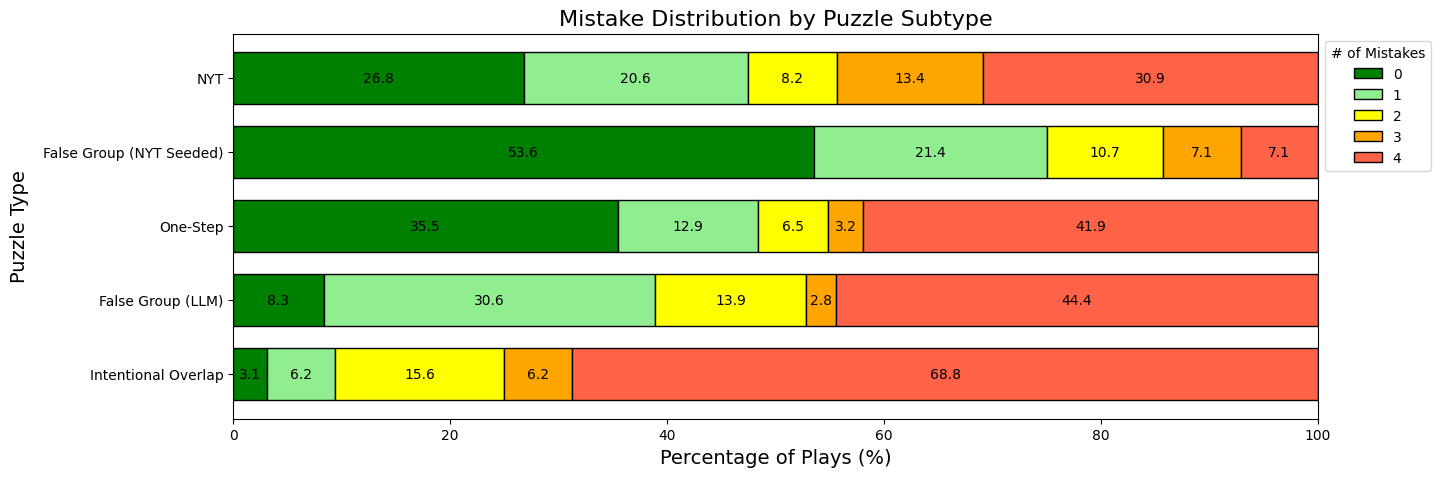} 
\caption{Number of mistakes made by percentage of puzzle plays, grouped by puzzle sub-type. AI-generated puzzles involving intentional overlaps proved most difficult to human players, while those seeded with false groups from existing NYT puzzles proved significantly easier than actual NYT puzzles.}
\label{fig:mistakes}
\end{figure*}

\begin{table}[t]
    \centering
    \caption{Solve rates of puzzles by type}
    
    \begin{tabular}{lc}
    \toprule
    \textbf{Type} &  \textbf{Solve rate} \\
    \midrule
    One-Step &  58.06\% \\
    Intentional Overlap & 31.25\% \\
    False group (LLM) &  55.56\% \\
    False group (seeded) &  92.86\%  \\
    All AI &  58.51\% \\
    NYT &  69.07\%\\
    \bottomrule
    \end{tabular}
    \label{tab:solve_rate}
\end{table}

We collect 78 responses from 52 unique users (by username), sharing links to the survey via social media. We encouraged participants to complete the survey multiple times, with the most completions by a single user being 6.

\subsection{AI vs Real Puzzles}
Figure \ref{preference_data} shows user preferences on AI generated puzzles versus real New York Times puzzles. We compare the percentage of user responses to questions 4-6 across all AI-generated puzzles, as well as by sub-type.
In roughly half of head-to-head comparisons against expert \Times{} puzzles, generated \Connections{} puzzles were judged to be equally or more enjoyable, creative, and difficult.

\subsection{One-step vs Iterative Puzzles}
One-step puzzles were largely rated as being less enjoyable and less creative than their \Times{} counterparts, but performed better in terms of difficulty. 
Our best performing group in terms of overall enjoyment (as indicated by responses to question 6) was LLM false group puzzles, beating \Times{} puzzles in $42.86\%$ of responses and tying in $14.29\%$. Users found intentional overlap puzzles the hardest, rating them more difficult than their NYT counterparts in $60\%$ of responses and equally difficult in $20\%$ of responses. This difficulty is reflected in playthrough data, where intentional overlap puzzles have the lowest overall solve rate. In terms of creativity (question 4), intentional overlap puzzles perform best, though LLM-generated false group puzzles are strictly preferred more often.

In general, difficulty is the category where generated puzzles appear to come closest to the \Times{}.

\paragraph{Solve rate}
We find a significant relationship between the puzzle sub-type and the solve rate, $X^2$(4, N=194) = $19.71$, $p<.01$. The difference in solve rate between \Times{} puzzles and all AI methods taken together is not statistically significant, though noticeable differences appear when investigating by AI puzzle sub-type. Intentional overlap puzzles were by far the most difficult for users, being solved in only $31.25\%$ of play sessions. On the contrary, seeded false group puzzles seem trivially easy, being solved in $92.86\%$ of play sessions. 

The mistake distribution of AI-generated puzzles (Figure \ref{fig:mistakes}) tells an interesting story. We find significantly different distributions within each sub-type. In intentional overlap puzzles, players made four mistakes in the majority of plays ($68.75\%$), and very rarely completed the puzzle with none. Conversely, players solved seeded false groups perfectly in the majority of plays ($53.57\%$), and very rarely failed the puzzle overall. 

\subsection{Difficulty}
We analyzed users' play data to determine whether our attempts to add interplay via false groups worked as we expected. For each false group puzzle, we calculated how often players made a false group guess: selecting any subset of the false group (2, 3, or 4 words). We found significant differences between the LLM- and NYT-seeded versions. For all LLM-seeded false group puzzles, players guessed a subset of the false group in half or more of the total play sessions. In one puzzle, every single attempt included one of these false group guesses.

For seeded false group puzzles, players weren't tricked as easily. While 2 puzzles elicited false group guesses, 3 puzzles had no false group guesses.

Players often used the free response section of the survey to provide their opinions on puzzle difficulty. Users were surprisingly perceptive in identifying the manner in which difficulty was introduced into generated puzzles. One user, playing an intentional overlap puzzle, writes ``\textit{There was a bit more overlap in words to categories}.'' Some users cite the difficulty as a reason they enjoyed the puzzled more: ``\textit{Too many word overlaps in the second puzzle, but it was more interesting and accurate to their categories}''.

Users were split on whether this ambiguity in groupings was a positive or negative. Six users specifically call out ambiguity and overlap in the AI puzzle as a negative, while seven cite it as a positive and reason for preferring the puzzle overall.

Additionally, despite not disclosing that AI was used as a source for any puzzles, multiple users wrote that they felt a puzzle was AI generated, though they weren't always accurate in their selection.

\section{Discussion}
\subsection{Difficulty discrepancy}
Seeded false groups stand as an outlier in our data---they are by far the easiest to solve. 
We hypothesize that this is due to the ``difficulty'' of the false groups. In particular, seeded false groups were drawn randomly from NYT puzzles, and therefore sometimes involved obscure connections. As a result, these groups were less likely to ``jump out'' at and distract players, whereas LLM-generated false groups were liable to be more generic, eye-catching, and therefore distracting, leading to more false group guesses.

We further note that LLM false groups were selected as the most similar set words in the candidate pool as measured by cosine similarity. The average similarity score for NYT-seeded false groups in our user study is indeed lower ($0.47$) than the average score for LLM-generated false groups ($0.52$). Further, two seeded groups have high similarity scores ($0.51$ and $0.60$), while the other three have much lower similarity ($< 0.43$). We note that the two groups with high similarity elicited false group guesses from users, and the three with low similarity elicited none.

Difficulty is the category in which our generated puzzles perform best against NYT puzzles. Intentional overlap puzzles were rated as more difficult in the majority of cases, similar to LLM false groups. This is likely due to the way we intentionally introduce the overlap or false connection difficulty into \textit{every} group in the puzzles. While NYT puzzles may incorporate just a few overlapping words, or small false connection groups, we try to maximize these types of difficulty in their respective puzzle types. This may lead to difficulty that exceeds the average NYT puzzle. Decreasing this difficulty, by removing the difficulty incentives from generated group prompts, may help align the difficulty with what players have come to expect from \Connections{}.

\subsection{Interpretation of user preferences}
We intentionally limit the number of questions in the survey to reduce user fatigue. One notable metric that we do not include in our survey relates to puzzle ``fairness''. A puzzle that uses uncommon vocabulary or illogical groupings may be much harder for players, but not particularly fair in its difficulty. Although we use the Puzzle Editor LLM to try to combat these issues, it is not perfect at correcting puzzles.

\begin{table}[h]
\adjustbox{max width=\columnwidth}{%
\centering
\begin{tabular}{|l|l|}
\hline
\textbf{Category} & \textbf{Words} \\
\hline
\textit{NBA TEAMS} & ``bucks", ``heat", ``jazz", ``nets" \\
\textit{SLANG FOR TOILET} & ``can", ``head", ``john", ``throne" \\
\textit{THINGS WITH WINGS} & ``airplane", ``angel", ``bird", ``pegasus" \\
\textit{NFL PLAYERS} & ``bear", ``bill", ``brown", ``commander" \\
\textit{\_\_\_ ROAD} & ``abbey", ``high", ``rocky", ``silk" \\
\hline
\end{tabular}}
\caption{\Connections{} groups from published puzzles used as false groups}
\label{table:seed_categories}
\end{table}

\subsection{\Connections{} as a PCG domain}
Generating convincing \Connections{} puzzle is a uniquely difficult domain. To generate a puzzle on-par with the original, a method must be capable of proposing unique and clever categories, on the same level as professional puzzle builders. \Connections{} is published by a single source---the \Times{}---and players have come to expect a distinct level of quality and challenge from their daily puzzles. Group themes can't be too obvious, but they also can't be ``reaches''. This necessitates an advanced semantic understanding, combined with a sufficient model of human logic.

On top of this, user feedback shows the importance of the unspoken yet understood ``principles'' underlying \Connections{}. While some constraints are easily verifiable (Unique Names), others are harder to measure (Varied Categories). When even one of these principles is broken, players feel frustrated and cheated rather than accomplished. 

Despite these challenges, a direct comparison to \NYT{} doesn't tell the whole story. Though many generated puzzles may not beat real puzzles in direct comparison, users still enjoy solving them, and in some cases prefer them. Though small flaws are common, we find some surprising ability for cleverness and wordplay by LLMs. With further refinement of prompts, particularly in distilling the abstract concepts that make \textit{Connections} so widely enjoyed, we believe this method could produce more consistent results.

\subsection{Game generation as design research}

While the perhaps most obvious use case for the method we are developing here is to actually generate Connections puzzles, one can also see this work as a contribution to game design research. The implemented generator can be seen as a test of a theory of how Connections puzzles are designed. The implemented theory is based on Wyna Liu's article describing her own puzzle creation methodology, and the success of the method in creating good Connections puzzles can be seen as corroboration of that theory. One might then iteratively refine the theory and the corresponding process to arrive at an even better theory of what makes for a good Connections puzzle.

\section{Limitations}

\subsection{Cost}
Our ability to perform ablation studies of various prompting techniques, as well as generate large quantities of puzzles was limited by computational budget. The GPT-4 API is significantly slower and more expensive than other models, such as \texttt{gpt-3.5-turbo} or open source LLMs such as Llama 3 or Mistral. Open-source LLMs can be run locally without cost, enabling more in-depth state evaluation, thought generation, and editing steps that are cost-prohibitive with paid APIs. However, preliminary experiments indicate they are not as capable at following the complicated system prompts we author for GPT-4.

\subsection{User study data}
We rely on user's self-reporting whether or not they have seen a particular \Connections{} puzzle before. This may not be 100\% accurate, leading to some bias if users were able to identify previous NYT puzzles. 
Our analysis and confidence in our conclusions are limited by the number of responses, as well as our methodology. \Connections{} puzzles are inherently challenging to parse, forcing players to process a scrambled grid of seemingly unrelated words. Comparing two puzzles side-by-side may not be an optimal way of measuring a puzzle's difficulty.

\section{Future Work}

While our experiments show that---somewhat remarkably---\Connections{} puzzles generated entirely by A.I. can be competitive with puzzles generated by human experts along several axes, perhaps a more interesting end goal is to explore how such generative pipelines could ultimately assist human designers.
Our method could easily be adapted to add to one or more groups or overarching group themes, provided by a human designer. We explore LLMs as puzzle editors, but not as evaluators or editors of human-designed puzzles. LLM agents acting in these roles can facilitate a designer's creative flow and amplify their expressive ability (e.g. by providing incremental twists or edits on a partially complete puzzle when a designer is stuck, or calling on the immense knowledge base encoded in LLMs to enhance their level of reference).

\section*{Ethical Considerations}

There is ongoing, high-profile legal debate around the intellectual property of LLM-generated outputs---whether such models can be trained on copyrighted text, and how to properly attribute credit to output from models trained on such material.
Because the model used here is not trained on any bona fide NYT \Connections{} puzzles, it is incapable of plagiarizing the copyrighted work of human designers by stealthily regurgitating or paraphrasing what it may have memorized during training. On the other hand, we do few-shot prompt the model with some examples of NYT \Connections{} puzzles for the sake of demonstration and inspiration. We do not find that the model borrows from these puzzles in any kind of conspicuous fashion, though users of our system should be careful to compare outputs against human puzzles injected into prompts.

\section{Conclusion}
In this work, we present a method of generating word puzzles for \NYT{}' \textit{Connections} game using Large Language Models. The results of our user study show that the puzzles generated by our method are often competitive in terms of player preferences, and comparable in terms of difficulty. We argue that, with careful, domain-specific prompting, LLMs (i.e. OpenAI's GPT-4) are capable of generating novel and challenging \Connections{} puzzles, and could ultimately serve as viable design assistants for \Connections{}, or other, similar lateral thinking games relying on semantic understanding and creative associations between words and concepts.

Our puzzle generation pipeline involves a novel iterative prompting approach that combines LLM-generated categories, words and distractors with a sentence embedding cosine similarity metric to dynamically select for group difficulty. We incorporate design principles from \Connections{} (namely, false groups and intentional overlap distractors) into our generation pipeline, and find that these lead to significant differences in puzzle difficulty and preference among human players. We additionally demonstrate the compatibility of our method with a human-in-the-loop approach, showing how user-provided ``false groups'' can be used to seed viable synthetic puzzles.

Still, generating entire \textit{Connections} puzzles remains a challenge requiring further study. The puzzle design principles incorporated into our method are far from exhaustive, and our user study is strictly an initial exploration: it stands to reason that repeated exposure to these synthetic puzzles could eventually lead to player fatigue, whereas human designers may adapt their approach in real-time to provide a dynamic experience for repeat players. Future work should seek to study the utility of such generative pipelines to puzzle designers themselves, and investigate ways in which the puzzle design principles integrated into the LLM prompting pipeline may be modified by designers to adapt to design goals that evolve over time.

\bigskip

\bibliography{aaai24}

\begin{thebibliography}{15}
\providecommand{\natexlab}[1]{#1}

\bibitem[{Aronow(2021)}]{Aronow_crossword}
Aronow, I. 2021.
\newblock Crossword Constructor Resource Guide.
\newblock \emph{The New York Times}.

\bibitem[{Aronow and Levine(2023)}]{Aronow_Levine_2023}
Aronow, I.; and Levine, E. 2023.
\newblock How to line up a great connections solve.
\newblock \emph{The New York Times}.

\bibitem[{Gallotta et~al.(2024)Gallotta, Todd, Zammit, Earle, Liapis, Togelius, and Yannakakis}]{gallotta2024large}
Gallotta, R.; Todd, G.; Zammit, M.; Earle, S.; Liapis, A.; Togelius, J.; and Yannakakis, G.~N. 2024.
\newblock Large language models and games: A survey and roadmap.
\newblock \emph{arXiv preprint arXiv:2402.18659}.

\bibitem[{Jaramillo et~al.(2020)Jaramillo, Charity, Canaan, and Togelius}]{Jaramillo_Charity_Canaan_Togelius_2020}
Jaramillo, C.; Charity, M.; Canaan, R.; and Togelius, J. 2020.
\newblock Word Autobots: Using Transformers for Word Association in the Game Codenames.
\newblock \emph{Proceedings of the AAAI Conference on Artificial Intelligence and Interactive Digital Entertainment}, 16(1): 231--237.

\bibitem[{Koo et~al.(2023)Koo, Lee, Raheja, Park, Kim, and Kang}]{koo2023benchmarking}
Koo, R.; Lee, M.; Raheja, V.; Park, J.~I.; Kim, Z.~M.; and Kang, D. 2023.
\newblock Benchmarking cognitive biases in large language models as evaluators.
\newblock \emph{arXiv preprint arXiv:2309.17012}.

\bibitem[{Liu(2023)}]{Liu_nyt}
Liu, W. 2023.
\newblock How our new game, connections, is put together.
\newblock \emph{The New York Times}.

\bibitem[{Long(2023)}]{long2023large}
Long, J. 2023.
\newblock Large Language Model Guided Tree-of-Thought.
\newblock \emph{arXiv preprint arXiv:2305.08291}.

\bibitem[{Raphael(2020)}]{raphael20brief}
Raphael, A. 2020.
\newblock A Brief History of Word Games.
\newblock \emph{The Paris Review}.

\bibitem[{Schulhoff et~al.(2024)Schulhoff, Ilie, Balepur, Kahadze, Liu, Si, Li, Gupta, Han, Schulhoff et~al.}]{schulhoff2024prompt}
Schulhoff, S.; Ilie, M.; Balepur, N.; Kahadze, K.; Liu, A.; Si, C.; Li, Y.; Gupta, A.; Han, H.; Schulhoff, S.; et~al. 2024.
\newblock The Prompt Report: A Systematic Survey of Prompting Techniques.
\newblock \emph{arXiv preprint arXiv:2406.06608}.

\bibitem[{Sudhakaran et~al.(2023)Sudhakaran, González-Duque, Glanois, Freiberger, Najarro, and Risi}]{sudhakaran2023prompt}
Sudhakaran, S.; González-Duque, M.; Glanois, C.; Freiberger, M.; Najarro, E.; and Risi, S. 2023.
\newblock MarioGPT: Open-Ended Text2Level Generation through Large Language Models.
\newblock arXiv:2302.05981.

\bibitem[{Todd et~al.(2023)Todd, Earle, Nasir, Green, and Togelius}]{todd2023level}
Todd, G.; Earle, S.; Nasir, M.~U.; Green, M.~C.; and Togelius, J. 2023.
\newblock Level Generation Through Large Language Models.
\newblock In \emph{Proceedings of the 18th International Conference on the Foundations of Digital Games}, 1--8.

\bibitem[{Todd et~al.(2024)Todd, Merino, Earle, and Togelius}]{todd2024missed}
Todd, G.; Merino, T.; Earle, S.; and Togelius, J. 2024.
\newblock Missed Connections: Lateral Thinking Puzzles for Large Language Models.
\newblock \emph{arXiv preprint arXiv:2404.11730}.

\bibitem[{Treutlein et~al.(2024)Treutlein, Choi, Betley, Anil, Marks, Grosse, and Evans}]{connectingdots}
Treutlein, J.; Choi, D.; Betley, J.; Anil, C.; Marks, S.; Grosse, R.~B.; and Evans, O. 2024.
\newblock Connecting the Dots: LLMs can Infer and Verbalize Latent Structure from Disparate Training Data.
\newblock \emph{arXiv preprint arXiv:2406.14546}.

\bibitem[{Wei et~al.(2022)Wei, Wang, Schuurmans, Bosma, Xia, Chi, Le, Zhou et~al.}]{wei2022chain}
Wei, J.; Wang, X.; Schuurmans, D.; Bosma, M.; Xia, F.; Chi, E.; Le, Q.~V.; Zhou, D.; et~al. 2022.
\newblock Chain-of-thought prompting elicits reasoning in large language models.
\newblock \emph{Advances in Neural Information Processing Systems}, 35: 24824--24837.

\bibitem[{Yao et~al.(2023)Yao, Yu, Zhao, Shafran, Griffiths, Cao, and Narasimhan}]{yao2023tree}
Yao, S.; Yu, D.; Zhao, J.; Shafran, I.; Griffiths, T.~L.; Cao, Y.; and Narasimhan, K. 2023.
\newblock Tree of thoughts: Deliberate problem solving with large language models.
\newblock \emph{arXiv preprint arXiv:2305.10601}.

\end{thebibliography}

\appendix

\end{document}